\pdfoutput=1

\documentclass[11pt]{article}

\usepackage[]{acl}
\usepackage{hyperref}

\usepackage{times}
\usepackage{latexsym}

\usepackage[T1]{fontenc}

\usepackage[utf8]{inputenc}

\usepackage{microtype}

\usepackage{{booktabs}}
\usepackage{multirow}
\usepackage{tcolorbox}
\usepackage{algorithmic}

\usepackage{array}
\usepackage{caption}
\usepackage{subcaption}
\usepackage{amsmath,amssymb,stmaryrd}
\usepackage{todonotes}
\usepackage{xcolor}
\usepackage{booktabs}

\usepackage{xspace,mfirstuc,tabulary}

\newcommand{\instr}

\newcommand{\vheart}{\ensuremath\spadesuit}
\newcommand{\heart}{\ensuremath\heartsuit}
\newcommand{\diamondsmall}{\ensuremath\diamondsuit}
\newcommand{\club}{\ensuremath\clubsuit}

\usepackage{pifont}
\newcommand{\xmark}{\ding{55}}%

\usepackage{amsthm,amsmath,amsfonts,bm,xspace}
\usepackage{color}

\newcommand{\system}[1]{\textsc{#1}}
\newcommand{\data}[1]{\textsc{#1}}
\newcommand{\metric}[1]{\textsc{#1}}
\newcommand{\deleting}{deleting\xspace}
\newcommand{\obscuring}{obscuring\xspace}
\newcommand{\Deleting}{Deleting\xspace}
\newcommand{\Obscuring}{Obscuring\xspace}

\newcommand{\ourBenchmark}{\data{NaP$^2$}\xspace}
\newcommand{\personaChat}{\data{PERSONA-CHAT}\xspace}

\newcommand{\privacyLeakage}{\metric{Privacy\_NLI}\xspace}
\newcommand{\invprivacyLeakage}{\metric{1-Privacy\_NLI}\xspace}
\newcommand{\mauve}{\metric{MAUVE}\xspace}
\newcommand{\rouge}{\metric{ROUGE-1}\xspace}
\newcommand{\rougeLsum}{\metric{ROUGE-Lsum}\xspace}

\newcommand{\pStrict}{\metric{SPrivacy}\xspace}
\newcommand{\rStrict}{\metric{SRel}\xspace}
\newcommand{\nStrict}{\metric{SNatural}\xspace}
\newcommand{\llmNatural}{\metric{LLM-Natural}\xspace}

\theoremstyle{definition}

\newcommand{\chatGPT}{\system{GPT-3.5 turbo}\xspace}
\newcommand{\llama}{\system{Llama2-13B}\xspace}
\newcommand{\gptf}{\system{GPT4}\xspace}
\newcommand{\tfive}{\system{T5}\xspace}
\newcommand{\tfiveZero}{\system{T5\_zeroshot}\xspace}
\newcommand{\llamaZero}{\system{Llama2-13B\_zeroshot}\xspace}

\newcommand{\tfivesmall}{\system{T5-Base}\xspace}
\newcommand{\llamaparaph}{\system{Llama-paraph}\xspace}
\newcommand{\roberta}{\system{RoBERTa}\xspace}
\newcommand{\dpprompt}{\system{DP-Prompt}\xspace}
\newcommand{\dpbart}{\system{DP-BART}\xspace}
\newcommand{\flair}{\system{FLAIR-SCRUBBING}\xspace}
\newcommand{\bart}{\system{BART}\xspace}
\newcommand{\deberta}{\system{DeBERTa}\xspace}

\newcommand{\dpnr}{DPNR\xspace}
\newcommand{\dpforward}{DP-Forward\xspace}
\newcommand{\dpforwardU}{DP-Forward-utility\xspace}
\newcommand{\dpforwardP}{DP-Forward-privacy\xspace}

\definecolor{mgreen}{rgb}{0,0.7,0}

\def\eqref#1{(\ref{#1})}

\def\1{\bm{1}}

\def\vp{{\bm{p}}}

\def\vx{{\bm{x}}}
\def\vy{{\bm{y}}}

\DeclareMathAlphabet{\mathsfit}{\encodingdefault}{\sfdefault}{m}{sl}
\SetMathAlphabet{\mathsfit}{bold}{\encodingdefault}{\sfdefault}{bx}{n}

\title{\ourBenchmark: A Benchmark for Naturalness and Privacy-Preserving \\ Text Rewriting by Learning from Human}

\author{Shuo Huang\textsuperscript{\rm \heart}, 
William MacLean\textsuperscript{\rm \heart},
Xiaoxi Kang\textsuperscript{\rm \heart},
\\
\textbf{Qiongkai Xu}\textsuperscript{\rm \vheart},
\textbf{Zhuang Li}\textsuperscript{\rm \diamondsmall},
\textbf{Xingliang Yuan}\textsuperscript{\rm \club}, 
\textbf{Gholamreza Haffari}\textsuperscript{\rm \heart}
\textbf{Lizhen Qu}\textsuperscript{\rm \heart}\footnotemark[1], 
\\
\textsuperscript{\rm \heart} Monash University, \textsuperscript{\rm \vheart}Macquarie University,\textsuperscript{\rm \diamondsmall}RMIT University \textsuperscript{\rm \club}University of Melbourne\\
\textsuperscript{\rm \heart}\{shuo.huang1, xiaoxi, lizhen.qu, gholamreza.haffari\}@monash.edu,
\textsuperscript{\rm \diamondsmall}zhuang.li@rmit.edu.au
\\\textsuperscript{\rm \vheart}qiongkai.xu@mq.edu.au,\\\textsuperscript{\rm \club}xingliang.yuan@unimelb.edu.au}

\date{}

\begin{document}
\maketitle

\footnotetext[1]{Corresponding author.}

\begin{abstract}

 The widespread use of cloud-based Large Language Models (LLMs) has heightened concerns over user privacy, as sensitive information may be inadvertently exposed during interactions with these services. To protect privacy before sending sensitive data to those models, we suggest sanitizing sensitive text using two common strategies used by humans: i) deleting sensitive expressions, and ii) obscuring sensitive details by abstracting them. To explore the issues and develop a tool for text rewriting, we curate the first corpus, coined \ourBenchmark, through both crowdsourcing and the use of large language models (LLMs). Compared to the prior works on anonymization, 
 the human-inspired approaches result in more natural rewrites and offer an improved balance between privacy protection and data utility, as demonstrated by our extensive experiments. Researchers interested in accessing the dataset are encouraged to contact the first or corresponding author via email.
\end{abstract}

\section{Introduction}
\label{sec:intro}

%

Data sharing and information dissemination between AI models are pivotal in the AI era, particularly since the emergence of Large language models (LLMs). The remarkable performance of LLMs benefits from a large amount of shared and publicly available data. However, it is still challenging to balance between data privacy and information utility when training and utilizing such LLMs~\cite{pan2020privacy} with massive amount of data. Users or applications often interact with commercial LLMs by directly inputting raw text. Such interactions can inadvertently expose sensitive data, such as personally identifiable information (PII), to untrusted service or LLM providers~\cite{utpala2023locally}.

Redaction and anonymization techniques are widely applied to remove PII from texts, but they suffer from three major drawbacks~\cite{sanchez2014utility}. First, after anonymization, mentions of PII are either redacted or replaced by their entity types so that processed texts become \textit{unnatural} as it breaks grammatical flow, coherence and semantic clarity of sentences. Downstream applications need to be adapted or fine-tuned to cope with such unnatural texts. Second, it is still possible to recover private attributes from PII scrubbed text via reasoning~\cite{mireshghallah2023can,staab2023beyond}. Third, the presence of blacked-out parts or entity types may raise the awareness of a document's sensitivity in front of potential attackers.  

\begin{table}[t]
\centering
\small
\resizebox{\linewidth}{!}{%
\begin{tabular}{ll}
\toprule
ORI: & I have two teenage boys. \\
     & I have been to Los Angeles \\
     & a few years ago. \\
PER: & I am a single mom of two boys. \\
\midrule
\textbf{Human Rewrite:} \\
DEL: & I have been to Los Angeles \\
     & a few years ago. \\
OBS: & I have some children. \\
     & I have been to Los Angeles \\
     & a few years ago. \\
\midrule
\textbf{\tfivesmall trained on \ourBenchmark:} \\
Output: & I have been to Los Angeles \\
        & a few years ago. \\
\midrule
\textbf{\flair:} \\
Output: & I have \texttt{<MASK>} teenage boys. \\
        & I have been to \texttt{<MASK>} \texttt{<MASK>}. \\
\midrule
\textbf{\dpprompt:} \\
$\epsilon$-10: & Junior \\
$\epsilon$-100: & I have two teenage boys. \\
               & I have been to Los Angeles \\
               & a few years ago. \\
\bottomrule
\end{tabular}%
}
\caption{An example of rewriting a text (ORI) using deleting (DEL) and obscuring (OBS) as strategies based on personal information (PER). Output shows the \tfivesmall model finetuned with \ourBenchmark. Also shown are results from \flair and \dpprompt using $\epsilon$-10 and $\epsilon$-100.}
\label{tab:tease}
\vspace{-4mm}
\end{table}

The recent work on text anonymization \cite{dou2024reducing} introduced the task of self-disclosure abstraction, which involves rephrasing sensitive information into less specific terms while preserving utility (e.g., "I'm 16F" to "I'm a teenage girl"). A user study showed that 82\% of participants responded positively to the system, underscoring its practical relevance. However, the study focuses exclusively on rewriting mentions of private attributes, and the accompanying corpus includes only annotated spans of private attributes \textit{without human-authored reference rewrites}, limiting its suitability for reference-based evaluation metrics.

Alternatively, differential privacy (DP) provides a theoretical privacy guarantee for data release or dissemination mechanisms~\cite{dwork2006differential}. Prior works sanitize texts by perturbing texts either at the word-level or the sentence-level~\cite{mattern2022limits,igamberdiev2023dp,Igamberdiev.2022.COLING}. In order to reach a high-level of privacy guarantee, substantial noise needs to be injected into texts or their representations so that information utility drops sharply and the meanings of texts are changed significantly (see Table.~\ref{tab:tease}). Therefore, determining an optimal trade-off between privacy and utility for data release remains an unresolved challenge.

To address limitations of prior methods, we propose a human-inspired text editing approach—drawing on \textit{deleting} and \textit{obscuring} strategies~\cite{strengers2020adhering}—to enhance the naturalness and utility of rewritten texts while ensuring privacy, aligning with the suppression and generalization principles of k-anonymity~\cite{sweeney2002achieving} originally developed for structured data. As shown in Table~\ref{tab:tease}, given an utterance involving personal information stated in a persona, the strategy \textit{\deleting} simply removes all words mentioning sensitive information from the utterance, while \textit{\obscuring} substitutes sensitive expressions for more abstract and general expressions. In our example, the user requires explicitly rewrite private information about "a single mom of two boys". \textit{\deleting} removes entire part about this information and remain other parts untouched. While \textit{\obscuring} obscures the information about "boys" and "teenager" to simply "children" which generalizes the information to be protected.
Both strategies aim to make rewritten texts as \textit{natural} as possible such that i) they do not raise the awareness of potential attackers that rewrites are sanitized; and ii) downstream applications can directly process such natural rewrites without fine-tuning their models for any unnatural parts of texts.

To evaluate \textit{strategy-specific} rewriting models, we construct the \textit{first} \underline{Na}turalness and \underline{P}rivacy \underline{P}reserving Rewriting corpus, coined \ourBenchmark, based on the open-domain dialogue corpus \personaChat~\cite{zhang2018personaChat}. Unlike prior work that focuses solely on private attributes, our corpus incorporates text-based personalized privacy profiles. Hence, detection of personal information cannot be formulated as a multi-class classification task. We recruit university students to manually rewrite 895 utterances involving personal information as the \textit{manual evaluation set}. 

To promote the development of \textit{diverse} open-source solutions for this task, we apply \gptf to generate 3900 synthetic examples as the \textit{synthetic training} set because \gptf demonstrates the best performance on \personaChat among all evaluated models. We also design multiple automatic and human evaluation metrics for this task, including a \textit{novel} privacy metric \privacyLeakage. It utilizes a Natural Language Inference (NLI) model~\cite{liu2019roberta} to determine if a rewrite entails a personal information or not. The extensive comparative studies between the models trained on our corpus and the state-of-the-art (SOTA) text sanitization methods demonstrate the underlying challenges and yield the following key findings:
\begin{itemize}
    \item The \tfivesmall model~\cite{raffel2020t5} trained on our corpus is able to achieve a fairly high privacy preservation indicated by a \privacyLeakage of 93.81\%. Its performance is even significantly superior than \gptf according to human evaluation using \deleting. In contrast, the competitive DP methods have a \privacyLeakage score lower than 62.14\%.
    \item The privacy metric \privacyLeakage aligns well with the human judgments by having a Spearman's ranking correlation of 0.70.
    \item \gptf generates synthetic rewrites with decent trade-off between privacy and utility based on human evaluation, better than \chatGPT and the evaluated open-source LLMs in the zero-shot setting. Incorporation of such synthetic data improves the \tfivesmall model trained on human curated data by 7\% in terms of privacy preservation. 
\end{itemize}

\section{Naturalness and Privacy-Preserving Rewriting}

\subsection{Problem Definition}

\paragraph{Task.} Given an utterance $\vx$ and a sentence $\vp$ describing personal information, the task of naturalness and privacy-preserving rewriting aims to map $\vx$ into a natural sentence $\vy$ such that $\vy \in \mathcal{Y}^n$ does not reveal the personal information in $\vp$ and maximally preserves the non-private content in $\vx$. We define a natural sentence as one that is grammatically correct, fluent, and does not contain any artifacts such as blacked-out words or special symbols indicating omitted sensitive information. The rewrite space $\mathcal{Y}^n$ contains only natural sentences with maximum sequence length of $n$. Compared with DP mechanisms that prevent privacy leakage during model training~\cite{abadi2016DPSGD}, this task focuses on privacy-preserving data publishing or privacy protection at inference time while uploading user query.

When sanitizing texts, humans often hide sensitive information by avoiding sensitive words or replacing them with more general or abstract expressions~\cite{strengers2020adhering}. We expect machines to adopt similar strategies:

\begin{itemize}
\item \textbf{Deleting}: removing words or phrases in $\vx$ that leak personal information specified in $\vp$;
\item \textbf{Obscuring}: replacing sensitive words or phrases in $\vx$ with more general or abstract expressions to avoid compromising privacy.
\end{itemize}

\begin{table*}[t]
\centering
\scriptsize
\setlength{\tabcolsep}{3pt}
\begin{tabular}{|l|c|p{1.9cm}|c|c|p{2.1cm}|c|c|}
\hline
\textbf{Dataset} & \textbf{Year} & \textbf{Source} & \textbf{Human} & \textbf{Synthetic} & \textbf{Rewrite Type} & \textbf{Form of Private info} & Size \\
\hline
\textbf{NAP2 (Ours)} & 2025 & PERSONACHAT & \checkmark &  \checkmark & Delete / Obscure & privacy profile & Small\\
Self-Disclosure~\cite{dou2024reducing} & 2024 & Reddit Post& \xmark & \checkmark & Obscure & text span & Small\\
SythPAI~\cite{yukhymenko2024synthetic} &2024 & Reddit sytle &    \xmark       &  \xmark   & \xmark    &    PII    &    Large    \\
Text Anon. Benchmark (TAB)~\cite{pilan2022text} & 2022 & ECHR legal cases & \xmark& \xmark & Masking & PII & Large \\
TextWash\cite{kleinberg2022textwash} &         2022 & Wikipedia Bio &    \xmark    &    \xmark   &  Masking     &   PII     & Medium  \\
\hline
\end{tabular}
\caption{Comparison of recent datasets for text anonymization or privacy-preserving rewriting. Human and Synthetic refers to if there is any generated rewrite for the dataset either from human or LLMs. Rewrite type includes deletion, obfuscation. Masking indicates the detection private entities is redacted as entity type}
\label{tab:anonymization_datasets}
\end{table*}
 
\paragraph{Corpus Overview.} Our corpus \ourBenchmark consists of a small manually curated dataset for both training and testing (Sec. \ref{sec:manual_corpus}), and a large synthetic dataset distilled from \chatGPT and \gptf for training data augmentation (Sec. \ref{sec:synthetic_data}). According to our evaluation stated below, human rewrites with \obscuring achieve the best trade-off between privacy and utility, and the naturalness of \gptf generated texts is on par with that of human rewrites.

\paragraph{Comparison with existing datasets}
Our dataset stands out from other recent anonymization datasets by providing both human and synthetic rewrites based on diversified privacy profiles, rather than simple PII spans. Unlike datasets such as Self-Disclosure, which rely on LLM-generated rewrites focused on detected text spans, \ourBenchmark introduces more explicit rewrite operations including deletion and obscuration. Additionally, while other datasets emphasize masking or detection, \ourBenchmark offers more natural rewrites grounded in persona-level privacy, the rewrite option are clearly stated as obscure and delete which can facilitate different privacy protection level.

\subsection{Manually Curated Corpus}
\label{sec:manual_corpus}
The corpus \personaChat associates each multi-turn chit-chat with two personas, each of which is a set of sentences describing the corresponding personality. Detailed information for \personaChat are displayed in the Appendix.~\ref{app:personachat} Hence, it is straightforward to measure if an utterance leaks personal information in the relevant persona. From another point of view, a persona can be regarded as a user-specific privacy profile, which states what information needs to be protected. For instance, one user might consider their marital status as sensitive information requiring privacy protection, while another user may not prioritize it.

The manual created evaluation set extends the test set of \personaChat with human-authored rewrites. As not all utterances reveal private information in personas, we apply the automatic alignment methods to pair an utterance involving personal information with the corresponding sentence in a persona.
Formally, given a dialogue $\mathcal{D}$, suppose there are $m$ utterances $\mathcal{X}_i=\{\vx_1,\vx_2,...,\vx_m\}$ associated with a persona $\mathcal{P}_i=\{\vp_1,\vp_2,...,\vp_n\}$, we aim to compute an alignment score $s_{ij}$ between $\vx_i \in \mathcal{X}_i$ and $\vp_j \in \mathcal{P}_i$ indicating to what degree $\vx_i$ leaks personal information in $\vp_j$. 

We formulate the computation of alignment scores as an NLI problem. Namely, if $\vx_i$ entails $\vp_j$, it is highly likely that $\vx_i$ leaks information in $\vp_j$. Specifically, we reuse the \roberta model trained on Multi-Genre Natural Language Inference (MNLI) corpus~\cite{mnli}, which is available from Huggingface, to compute the probability of $p(y = \text{entail} | \vx_i, \vp_j)$ as $s_{ij}$. We find out that this simple approach significantly outperforms \system{Sparse-Max} and \system{Sharp-Max} proposed in~\cite{xu2020privacyDetection} on a random sample of 200 ground-truth pairs. We manually check the candidates among the pairs with a score higher than a threshold and keep only the well aligned ones.

For each selected sentence-persona pair, we recruit annotators from Amazon Mechanical Turk (AMT) to rewrite utterances w.r.t. the aligned persona sentences using both \Deleting and \Obscuring.

In our preliminary experiments, we observe that even though annotators endeavor to generate decent rewrites, many of them could not clearly identify and strictly stick to the required strategies. Therefore, we prepare a small sample of pairs as a pre-test to select qualified annotators. In addition, we employ a rigorous procedure for quality check. We wrap up 15 sentence-persona pairs as a batch and ask annotators to rewrite them using the required strategies. Then, we manually check the rewritten batches, we only accept those that are written using the required strategy. The averaged acceptance rate of the rewrites is 47.97\%, demonstrating the challenge of collecting a high-quality rewriting dataset with specific rewriting requirements. As a result, we collect 895 pairs annotated with one rewrite per strategy. We further split this corpus into a cross-validation (CV) set, a validation and a hold-out test set with 655, 140 and 100 instances, respectively.

\paragraph{Data Statistics.}
We analyze the manually curated corpus using averaged word length in sentences (Len.) and distinct unigrams divided by the total number of words (Dist.)~\cite{li2016diversity}. The statistics of the dataset is given in Table~\ref{tab:corpus_statistics}. \Deleting tends to produce more concise rewrites, while \obscuring is slightly longer than ORIGINAL sentences. Although the average length increases, the diversity score for \obscuring is still ascending, compared with original sentences. This shows the high diversity of word usage using \obscuring.

%
\begin{table}[t]
\centering
\footnotesize
\begin{tabular}{l rr rr rr}
\toprule
& \multicolumn{2}{c}{CV} & \multicolumn{2}{c}{Valid} & \multicolumn{2}{c}{Test}\\
& Len. & Dist. & Len. & Dist. & Len. & Dist. \\
\midrule
ORI & 13.7 & 0.148 & 13.6 & 0.257 & 13.5 & 0.248\\ 
\midrule
DEL & 8.0 & 0.190 & 8.4 & 0.298 & 8.5 & 0.279\\
OBS & 14.1 & 0.160 & 13.9 & 0.266 & 14.3 & 0.250\\
\bottomrule
\end{tabular}
\caption{Statistics of original sentence (ORI), rewrites with \deleting (DEL) and \obscuring (OBS) on the CV set, validation and test set of the manually curated dataset, using average length (Len.) and distinct token (Dist.)}
\label{tab:corpus_statistics}
\vspace{-2mm}
\end{table}

\subsection{Synthetic Data Augmentation}
\label{sec:synthetic_data}
We employ the \roberta NLI model to align utterances with persona sentences in the training set of \personaChat and keep only the pairs with an entailment probability above 0.3. This threshold leads to high recall low precision alignments so that \gptf is employed to check if there is indeed a privacy leakage. Among them, we randomly sample 3900 pairs to generate synthetic rewrites by using \gptf. The resulting dataset is used to augment the training set of the manually created corpus to mitigate the data scarcity issue.

Prior studies show that \gptf is one of the strongest few-shot learner~\cite{brown2020gpt3}. Therefore, we carefully design prompts and in-context examples to use it for privacy-aware rewriting. Given an utterance-persona pair, we use the following prompt for a selected rewriting strategy.

\begin{tcolorbox}
Rewrite this sentence, \textless deleting / obscuring\textgreater{} any private information.

Example rewrites are:

\textless{}$\text{IN-CONTEXT\_EXAMPLES}$\textgreater{}

Only return the rewritten sentence, nothing else.

Private information present is: [\$PERSONA]. \\
The sentence to rewrite is: [\$UTTERANCE].
\end{tcolorbox}

where \$X denotes a placeholder for the corresponding information. The $k$ in-context examples are selected from a combination of the validation set of the manually curated corpus and a set of non-sensitive utterances which do not leak personal information. Each of the in-context examples in the validation set contains an utterance, a persona sentence, and a human rewrite using the given strategy, while an example from the non-sensitive set includes only an utterance. The in-context examples are found by $k$-nearest neighbour search using the sentence embeddings of utterances~\cite{reimers2019sentence}. In this work, given an utterance, we select the top-$1$ most similar example from the validation set and one example from the non-sensitive set. The latter is used to instruct \gptf that it should not rewrite an utterance if there is no privacy leakage detected.

\subsection{Human Evaluation}
\label{sec:human_eval}
 Three university students are recruited to check their quality on a set of 100 instances sampled from the test set of the manual corpus. Hence, an utterance-persona pair in the sample includes a human rewrite, a rewrite from \chatGPT and \gptf respectively. For each rewrite, a student is instructed to answer the following questions from the perspectives of privacy leakage (Q1), semantic relevance (Q2) and naturalness (Q3) which is detailed in Appendix \ref{app:question design}. Each question is answered by three university students. To deal with possible disagreements, we take the \textit{majority vote} as the final answer. For annotation, the three annotators achieved Fleiss' Kappa\cite{falotico2015fleiss} inter-annotator agreement score with 0.47 which is acceptable for classification problem.

In order to use a score to summarize the performance w.r.t. each criteria, we calculate the percentage of choosing the option (a) as the majority vote for each question above on the human evaluation test set, referred to as \pStrict, \rStrict, and \nStrict. They indicate the percentage of rewrites having no privacy leakage, complete semantic relevance, full naturalness, respectively. 

\begin{table}[]
\resizebox{0.95\columnwidth}{!}{%
\begin{tabular}{c|c|c|c}
\hline
\multicolumn{1}{l|}{} & \pStrict & \rStrict & \nStrict  \\ \hline \hline
Human\_deleting  & 82.00\%        & 76.00\% & 95.00\% \\ \hline
GPT3.5\_deleting   & 34.00\%        & 94.00\% & 72.00\% \\ \hline
GPT4\_deleting   & 49.00\%        & 92.00\% & 99.00\% \\ \hline \hline
Human\_obscuring & 81.00\%        & 97.00\% & 98.00\% \\ \hline
GPT3.5\_obscuring  & 61.00\%        & 90.00\% & 95.00\% \\ \hline
GPT4\_obscuring  & 66.00\%        & 95.00\% & 99.00\% \\ \hline
\end{tabular}%
}
\caption{Comparison between \chatGPT, \gptf, and human rewrites.}
\label{tab:synthetic_data_quality}
\end{table}
\begin{figure}[htb]
    \centering
    \includegraphics[width=\columnwidth]{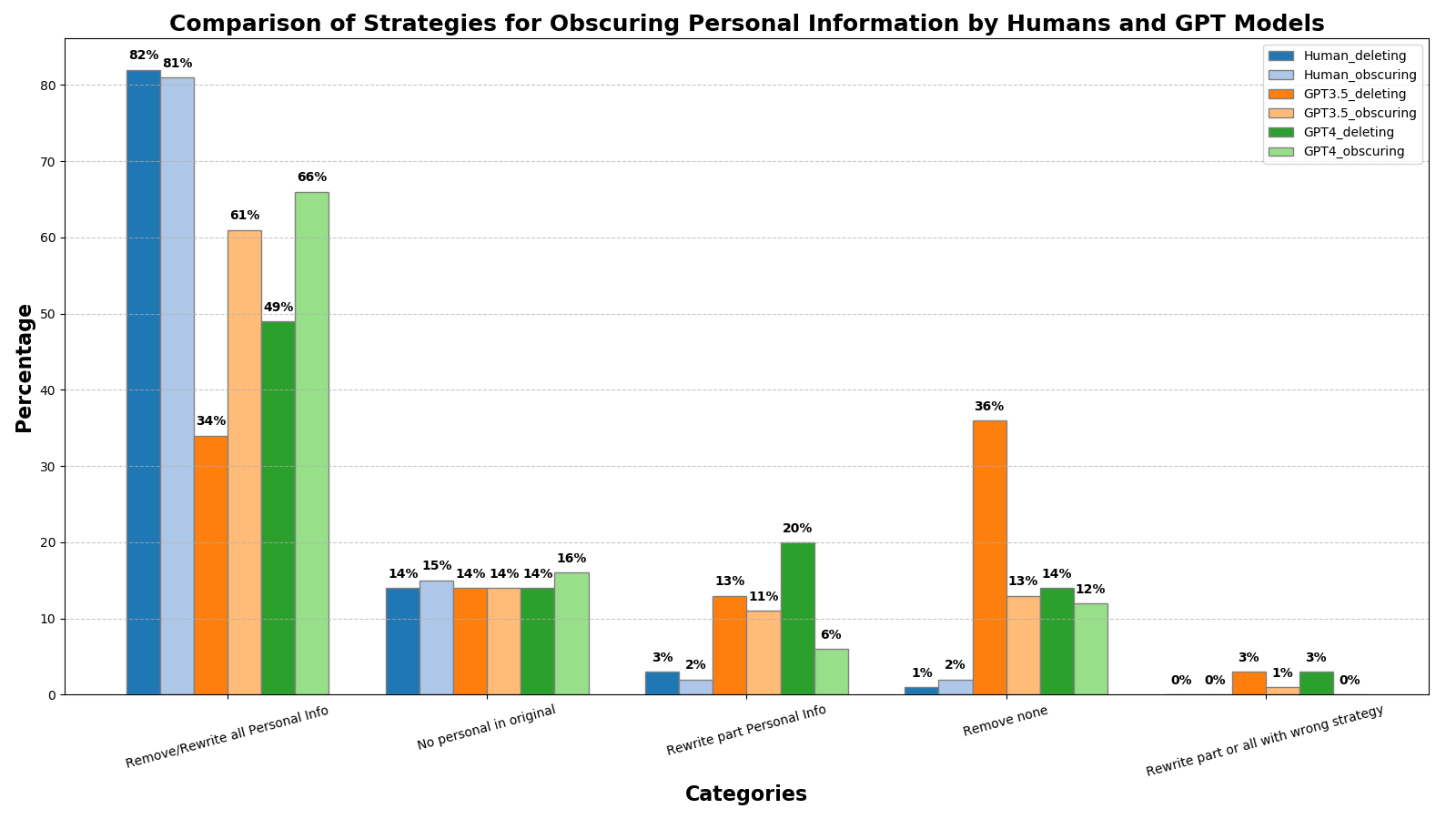}
    \caption{Human evaluation of privacy leakage.}
    \label{fig:human_synthetic}
\end{figure}
To understand the quality of rewrites in our corpus, we compare \gptf outputs with those of \chatGPT using the same prompts, as well as with human rewrites. The key results are summarized in Table \ref{tab:synthetic_data_quality}. Human rewrites achieve the highest level of privacy protection with both strategies, outperform the best rewriting model \gptf by at least 15\%. Human rewrites with \obscuring achieve the best balance between privacy and utility in comparison with alternative methods. Both OpenAI models completely preserve personal information in over 60\% of utterances by using \obscuring, but struggle to implement the \deleting strategy for the same purpose. A close investigation on the percentages of individual Q1 answer in Fig. \ref{fig:human_synthetic} demonstrates that both models fail to delete private expressions completely in over 34\% of the utterances involving sensitive information. \chatGPT is significantly worse than \gptf in terms of sanitization. Only 
a small proportion of the errors are attributed to applying an incorrect strategy.      

\section{Experiments}
\subsection{Rewriting Models}
\label{sec:models}
In this section, we establish a baseline approach to assess the efficacy of existing privacy protection solutions in removing private content within textual messages. This evaluation is pivotal to addressing the critical question: "\textbf{Are current privacy protection solutions adequately equipped to conceal privacy-sensitive content in utterances?}"
We compare the SOTA privacy rewriting models

To have a comprehensive comparison of privacy-preserving methods during model inference time, we compare several state-of-the-art privacy-preserving rewriting models, including \dpnr~\cite{lyu2020differentially}, \dpforward~\cite{du2023dpforward}, \llamaparaph, \dpprompt, \flair, and \dpbart. \dpnr applies Laplace noise to word representations, while \dpforward perturbs embedding matrices for sentence-level LDP detailed explaination for each baseline method can be foud in Appendix.~\ref{app:experiment}.

To assess the rewriting ability of models in zero shot, we consider LLMs fine-tuned on our corpus and apply the same prompts to the same pre-trained LLMs without any additional training. Specifically, we consider \tfivesmall, \llama, \chatGPT, and \gptf, and apply the prompt template introduced in Sec. \ref{sec:synthetic_data}. To distinguish them from the fine-tuned models, \tfivesmall and \llama in the zero-shot setting are referred to as \tfiveZero and \llamaZero, respectively.
In order to test our dataset and synthetic data augmentation, we consider \tfivesmall fine-tuned on our corpus, with or without synthetic data augmentation. We further use \bart to fine-tune our dataset to emphasize the effectiveness of our datasets. In addition, we assess \tfive-\ourBenchmark fine-tuned with and without DP-SGD~\cite{abadi2016DPSGD} to evaluate the impact of differential privacy on model performance.
\begin{table}[t]
\centering
\resizebox{\columnwidth}{!}{
\begin{tabular}{lrrrr}
\toprule
Method & \privacyLeakage & \pStrict & \rouge & \rougeLsum  \\
\midrule
\dpnr                      & 62.14\% & 25.00\% & 92.79\% & 92.79\% \\
\dpforward                 & 36.42\% & 0.00\%  & 99.91\% & 99.91\% \\
\dpprompt                 & 62.86\% & 0.00\%  & 42.18\% & 41.89\% \\
\dpbart                   & 78.22\% & 1.00\%  & 44.01\% & 43.15\% \\
\flair                    & 56.43\% & 0.00\%  & 67.75\% & 67.89\% \\
\tfiveZero-\deleting      & 70.00\% & 10.00\% & 16.62\% & 12.61\% \\
\tfiveZero-\obscuring     & 45.00\% & 45.00\% & 29.58\% & 23.80\% \\
\llamaZero-\obscuring     & 79.28\% & 16.00\% & 40.86\% & 40.12\% \\
\llamaZero-\deleting      & 77.14\% & 14.00\% & 68.28\% & 67.53\% \\
\llamaparaph-\obscuring   & 82.86\% & 31.00\% & 21.72\% & 20.05\% \\
\llamaparaph-\deleting    & 76.42\% & 16.00\% & 56.29\% & 54.91\% \\
GPT-3.5-\obscuring        & 87.14\% & 61.00\% & 66.66\% & 65.76\% \\
GPT-3.5-\deleting         & 74.29\% & 34.00\% & 69.13\% & 68.48\% \\
GPT-4-\obscuring          & 92.14\% & 66.00\% & 73.24\% & 72.63\% \\
GPT-4-\deleting           & 90.00\% & 49.00\% & 77.48\% & 77.08\% \\
\tfive-\ourBenchmark-\gptf & \textbf{93.81\%} & \textbf{72.00\%} & 73.01\% & 72.78\% \\
\bottomrule
\end{tabular}}
\caption{Evaluation and comparison of baseline methods.}
\label{tab:baseline}
\end{table}

\subsection{Evaluation Details}
\label{sec:metrics}
Prior studies focus on protect data privacy from membership inference attacks, reconstruction attacks, and sensitive attribute attacks etc.~\cite{mattern2022limits}. However, almost all of them focus on privacy preservation at the training time. In contrast, our target task is concerned with i) if a rewrite reveals personal information in a given persona, ii) preservation of non-sensitive content, and iii) naturalness of rewrites. Compared with the prior studies based on DP mechanisms, our setting is more close to that of natural language generation (NLG) tasks. Therefore, we evaluate the outcomes of the rewriting models by using NLG motivated automatic and human evaluation. 

For human evaluation, we use the same questionnaires and the metrics introduced in Sec. \ref{sec:human_eval} and ask annotators to answer each question in order to obtain the majority votes.

For all experiments involving model fine-tuning, we conduct five folds cross validation (CV) on the CV set of the manually curated corpus. In order to understand the usefulness of synthetic data, we also conduct experiments with the same models that augment the training set in each fold with 3,900 synthetic instances generated by \gptf.
\subsubsection{Automatic Evaluation Metrics.}
\textbf{Privacy Leakage.} We propose a novel metric, called \privacyLeakage, by using the \roberta model trained on the MNLI corpus, to infer to what degree it is possible to infer personal information in personas. As the NLI model classifies a pair of input texts into \textit{entailed}, \textit{contradicted}, or \textit{neutral}, we adopt $P(\text{entailed}|\vx, \vp)$ as the score of $privacy\_leakage$, e. Hence, we consider \privacyLeakage as 1- $privacy\_leakage$, denoting the privacy preserved by our method. The higher the metric, the more private information is preserved.
To validate the effectiveness of this metric, we incorporated recent NLI models trained with variant MNLI corpus to show the consistency.
Besides we also considered Sparse-MAX and Sharp-MAX as soft alignment score\cite{xu2020privacyDetection}, the detailed comparsion can be found in Appendix.~\ref{sparse_align}.

\textbf{Semantic Relevance.}
For assessing the preservation of semantic content, we consider \rouge and \rougeLsum~\cite{lin2004rouge} to compare generated rewrites with the corresponding references detailed introduction can be found in Appendix.~\ref{app:experiment}.




\subsection{Results and Discussions}


\begin{table}[]
\resizebox{0.95\columnwidth}{!}{%
\begin{tabular}{c|c|c|c}
\hline
\multicolumn{1}{l|}{} & \pStrict & \rStrict & \nStrict  \\ \hline \hline
Human\_deleting          & \textbf{82.00\%} & 76.00\%      & 95.00\% \\ \hline
\llama\_deleting          & 54.00\% & 49.00\%      & 87.00\% \\ \hline
\tfive-\ourBenchmark-\gptf\_deleting &    \textbf{72.00\%}   &  91.00\%    &   95.00 \% \\ \hline
\dpnr           & 1.00\%  & 0.00\%       & 19.00\% \\ \hline
Human\_obscuring &    \textbf{81.00\%}     &  97.00\%      &   98.00\% \\ \hline
\dpprompt           & 0.00\%  &1.00 \%       & 0.00\% \\ \hline
\dpbart           & 1.00\%  & 10.00\%       & 2.00\% \\ \hline
\flair           & 0.00\%  & 1.00\%       & 0.00\% \\ \hline

\llama\_obscuring         & 12.00\% & 14.00\%      & 86.00\% \\ \hline
\tfive-\ourBenchmark-\gptf\_obscuring &      \textbf{53.00\%}    &     93.00\%     &  98.00\%     \\

\bottomrule

\end{tabular}%
}
\caption{Human evaluation of the SOTA models.}
\label{tab:RQ1_result}
\end{table}

\begin{table}[]
\resizebox{0.95\columnwidth}{!}{%
\begin{tabular}{c|c|c}

\hline
\multicolumn{1}{l|}{} &  \nStrict  &     \llmNatural       \\ \hline \hline
Human\_deleting          &        95.00\% &        4.14        \\ \hline
\llama\_deleting          &     87.00\% &           4.67        \\ \hline
\tfive-\ourBenchmark-\gptf\_deleting &        95.00 \%  &  4.44  \\               \hline
\dpnr           &        19.00\%    &       2.01              \\ \hline
Human\_obscuring &           98.00\% &        4.37           \\ \hline
\dpprompt           &    0.00\%  &          1.14      \\ \hline
\dpbart           &  2.00\%  &          1.71        \\ \hline
\flair           & 0.00\%  &             3.05     \\ \hline

\llama\_obscuring         &  86.00\%         &    4.85       \\ \hline
\tfive-\ourBenchmark-\gptf\_obscuring &        98.00\%   & 4.36    \\

\bottomrule

\end{tabular}%
}
\caption{LLM as Naturalness Judge compared with Human preference}
\label{tab:llm_human_alignment}
\end{table}

\textbf{Efficacy of \ourBenchmark.} 
Table~\ref{tab:baseline} reports the evaluation of all methods. \tfivesmall fine tuned on the human rewrites and the synthetic data using both strategies outperform the DP based methods and zero-shot LLMs by a wide margin. \dpnr preserves more privacy than \dpforward, but results in a dramatic drop of information utility. The generated texts often have completely different meanings and have substantial grammatical errors, though some of them are still fluent. In contrast, \dpforward mostly copies inputs to outputs but rarely hide sensitive information. \llamaparaph produces frequently irrelevant texts, hence have fairly low \rouge and \rougeLsum scores. Besides, for convention personally identifiable information scrubbing method \flair, it can not effectively remove the private information in open-ended domain, only 40.71\% examples are successfully removing PII tokens. For \dpprompt and \dpbart, even \privacyLeakage are outperformed than other baseline models, the paraphrasing impairs the semantic of original sentence leading to low \rouge score.

We further investigate the rewriting quality w.r.t. each strategy based on human evaluation. We use the \tfivesmall model trained on the human rewrites and the synthetic data with both strategies, and apply it on the hold-out test set of each strategy. Table \ref{tab:RQ1_result} shows that the \tfivesmall model achieves superior performance over the baselines with both strategies. The naturalness of all generated rewrites is on par with that of human rewrites. Both zero-shot \llama models perform better than the best DP method \dpnr, which mostly perturbs non-sensitive contents or yields repeated words. The near-zero SPRIVACY scores observed for these methods stem from the nature of noise injection in embeddings or numeric representations. This results in two extremes: either no change in output due to insufficient perturbation or complete distortion of the generated output. The overall results are encouraging for a wide range of applications on edge devices, because our corpus is not huge and \tfivesmall contains only a few million parameters, which is a few hundred times smaller than \llama, \chatGPT and \gptf.

\textbf{Alignments between Automatic metrics and Human Evaluation.} We compare the ranking using \privacyLeakage with the corresponding human judgments in Table \ref{tab:baseline}. 
\tfive-\ourBenchmark-\gptf obtains the highest \invprivacyLeakage of 93.81\% in automatic evaluation, matching the highest \pStrict with 72.00\%. The results are aligned well among the rewriting models using the \obscuring. However, \privacyLeakage does not rank all rewriting models using \deleting in the same manner as humans. To quantify the alignments, we calculate a Spearman's correlation of 0.70 between \privacyLeakage and \pStrict among all models to demonstrate \privacyLeakage. The correlation between the models using \obscuring reaches even 0.83. 

\textbf{Naturalness Assessment via LLM-as-judge}
We also consider LLM as judge to score the naturalness of generated sentence. Specifically, we reuse the question from human questionnaire about naturalness and convert it to prompt template with score scale from 1-5. The prompt can be found in the Appendix~\ref{append:natural_template} The results in Table~\ref{tab:llm_human_alignment} demonstrate a generally strong alignment between \textbf{LLM-judged naturalness} and \textbf{human preference} (\nStrict), particularly for high-performing models such as \llama\_deleting, \tfive-\ourBenchmark-\gptf\_deleting, and \tfive-\ourBenchmark-\gptf\_obscuring, where LLM scores ($\geq$ 4.36) closely reflect human-rated naturalness ($\geq$ 95\%). However, notable discrepancies emerge for low-performing systems like \dpprompt and \flair, where LLMs assign moderately high naturalness scores (e.g., 3.05 for \flair) despite near-zero human ratings. This indicates that while LLMs can approximate human judgments in many cases, they may overestimate the fluency or coherence of outputs that humans find unnatural, underscoring the need for further calibration of LLM-based evaluation frameworks.





\textbf{Usefulness of the Synthetic Data.}
Table \ref{tab:RQ3_result} shows the result of using synthetic data for training rewriting models. We compare two different strategies: \deleting and \obscuring. The results shows that the model performs better with the synthetic data for both tasks. In particular, the model preserves more non-personal information compared to human rewrites in the \deleting task. With the synthetic data training the models, the model performance is 7\% better than the non-synthetic data model in terms of \deleting. The biggest gain of the synthetic data is obtained for improving the privacy protection of the rewriting model using \obscuring. 

\begin{table}[]
\resizebox{0.95\columnwidth}{!}{%
\begin{tabular}{c|c|c|c}
\hline
\multicolumn{1}{l|}{} & \pStrict & \rStrict & \nStrict  \\ \hline \hline
Human\_deleting          & \textbf{82.00\%}        & 76.00\% & 95.00\% \\ \hline
\tfive-\ourBenchmark-\gptf\_deleting            & 72.00\%        & 91.00\% & 95.00\% \\ \hline
non-Syn\_deleting        & 65.00\%        & 92.00\% & 93.00\% \\ \hline \hline
Human\_obscuring         & \textbf{81.00\%}        & 97.00\% & 98.00\% \\ \hline
\tfive-\ourBenchmark-\gptf\_obscuring           & 53.00\%        & 93.00\% & 98.00\% \\ \hline
non-Syn\_obscuring       & 4.00\%         & 92.00\% & 93.00\% \\ \hline
\end{tabular}%
}
\caption{Human evaluation results with and without synthetic data.}
\label{tab:RQ3_result}
\end{table}

\textbf{NLI score with different backbones}
\begin{table}[t]
\centering
\resizebox{\columnwidth}{!}{%
\begin{tabular}{llll}
\toprule
\privacyLeakage  & Mixed corpus & MNLI    & \pStrict  \\
\midrule\midrule
\dpnr                   & 80.54\%      & 53.69\% & 25\%              \\
\dpforward            & 68.87\%      & 48.06\% & 0\%               \\
\dpprompt             & 71.85\%      & 82.76\% & 0\%               \\
\dpbart               & 85.87\%      & 80.72\% & 1\%               \\
\flair                  & 62.16\%      & 61.49\% & 0\%               \\
\tfiveZero-\deleting  & 61.33\%      & 49.14\% & 10\%              \\
\tfiveZero-\obscuring & 61.33\%      & 19.26\% & 45\%              \\
\llamaZero-\obscuring         & 81.85\%      & 45.23\% & 16\%              \\
\llamaZero-\deleting       & 87.04\%      & 76.68\% & 14\%              \\
\llamaparaph-\deleting    & 79.42\%      & 65.91\% & 31\%              \\
\llamaparaph-\obscuring   & 82.69\%      & 52.21\% & 16\%              \\
GPT-3.5-\obscuring          & 90.33\%      & 53.51\% & 61\%              \\
GPT-3.5-\deleting         & 61.33\%      & 49.60\% & 34\%              \\
GPT-4-\obscuring          & 95.36\%      & 56.50\% & 66\%              \\
GPT-4-\deleting         & 89.26\%      & 47.59\% & 49\%              \\
Human Rewrite          & 97.01\%      & 63.20\% & 82\%              \\
\bottomrule\bottomrule
\end{tabular}%
}
\caption{\privacyLeakage score with \deberta as backbone finetuned with mixed entailment corpus and MNLI}
\label{tab:deberta}
\end{table}
In this section, we present the \privacyLeakage scores obtained using the \deberta model as the backbone, fine-tuned with different corpora, including a mixed entailment corpus and MNLI, to evaluate alignment with \pStrict, which serves as our human evaluation metric. As shown in Table \ref{tab:deberta}, the \deberta model achieves varying levels of performance across datasets. Models trained on the mixed corpus, such as \dpbart and \-deleting, achieve \privacyLeakage scores of 85.87\% and 95.36\%, respectively, with the former reaching 1\% and the latter 66\% alignment with \pStrict. Notably, human rewrites achieve a \privacyLeakage score of 97.01\%, with an alignment of 82\% with \pStrict. This comparison underscores the capability of different backbones and training strategies to achieve results close to human-level performance about privacy preservation while maintaining alignment with human evaluation metrics.

\section{Related Work}

The field of controllable text style transfer focuses on modifying specific attributes in texts, such as formality~\cite{briakou2021ola} and sentiment~\cite{li2018delete,li2022variational} while preserving the core semantic content. The advancement of text rewriting tasks is heavily dependent on the availability of high-quality corpora to assess generation quality. ~\citet{rao2018dear} collected a large-scale corpus GYAFC for initiating the research of formality style transfer to rewrite formal language. As for our task sensitive to privacy, which demands sophisticated alignment in rewriting utterances, the construction of a specialized corpus for high-quality privacy-sensitive rewrites are crucial.

There is a growing interest in protecting user privacy~\cite{chen2020fine,tigunova2019listening,xu-etal-2019-privacy,bevendorff2019heuristic} in NLP tasks. One way of protecting privacy is to implicitly remove the information in decision models, for example perturbing the representations via adversarial training ~\cite{li2018towards,elazar2018adversarial,barrett2019adversarial} or differential privacy~\cite{fernandes2019generalised,bo2019er}. In text rewriting which is close to our rewriting approach, local differential privacy are recently adapted to protect the data by adding customized noise~\cite{igamberdiev2022dp,igamberdiev2023dp}. Such adaptations in rewriting system mitigate the privacy leakage risk of original input however result in complete semantic change of inputs as the noise is independently drawn from the data and task. We consider a more generalised rewriting setting where the naturalness and general meaning of sentence are preserved.

Another series of work suggested to generate new sentences with less sensitive information~\cite{emmery2018style,xu-etal-2019-privacy}. Recent work has explored prompting LLMs to rewrite sentences containing private information, aiming to obscure sensitive content~\cite{emmery2018style,xu-etal-2019-privacy,staab2024large}. However, these approaches often rely on LLMs’ internal knowledge and struggle to align with nuanced human privacy expectations~\cite{dou2024reducing}. However in these works, the author does not control how private information to be rewritten in explicit way which may weaken the control for required privacy and human preference. Even advanced techniques like self-disclosure abstraction or adversarial anonymization face challenges in achieving robust, user-aligned privacy and often depend on powerful cloud-based models. In contrast, our work study private rewrite via more diversified free text and supports two rewriting strategies, offering a more flexible and general setting~\cite{strengers2020adhering}.








\section{Conclusion}
We introduce the task of naturalness and privacy-preserving text rewriting and collect a corpus \ourBenchmark based on \personaChat. The fundamental concept involves training models to learn human strategies, namely \deleting and \obscuring, for inference-time privacy. The \tfivesmall model trained on our corpus outperforms competitive zero-shot LLMs and DP methods by a wide margin. This work paves the way for future research on LLM-based rewriting techniques with a new focus on naturalness preservation.

\section*{Ethical Statement}
In this paper, we align our research practices with the principles outlined in the ACL Code of Ethics, fully endorsing its values. Our investigation has been conducted in compliance with these ethical standards. \par
The creation and assessment of \ourBenchmark have been conducted with a keen awareness of ethical considerations, especially regarding the involvement of human annotators. The necessity for human-annotated data to train conditional independence classifiers in our method is recognized as demanding significant effort. We have taken careful measures to ensure that this process is ethically sound, honoring the annotators' contributions by respecting their time and providing equitable compensation. \par
Moreover, the central objective of \ourBenchmark is to assess the relevance of generated responses in relation to their persona information and the difference between human evaluation and proposed automated metrics. The system is engineered to assign scores on a continuous scale from 0 to 1, with higher scores denoting greater relevance. It is designed to yield only these scores, without generating any information that could be deemed harmful or violate privacy.
\section*{Limitation}

Due to budgetary constraints associated with this project, we were unable to engage a vast number of annotators to rewrite the extensive dialogue datasets with respective rewrite strategies. Consequently, \ourBenchmark we compiled is somewhat limited in scope. While \ourBenchmark possesses sufficient volume to validate the core assertions of our study, it might not fulfill the expansive needs of commercial deployments. Industrial entities interested in utilizing our dataset could potentially address this limitation by adopting prompt tuning techniques or employing additional annotators to expand the dataset in accordance with our outlined methodology.\par

Our evaluation metric is specifically designed to assess the relevance of the generated responses. Although it demonstrates superior performance over baseline metrics in terms of privacy preservation and naturalness, the advantage it presents in relevance and specificity is less pronounced. Therefore, the development of innovative metrics tailored to specific evaluation criteria presents a valuable avenue for our future research endeavors.

\bibliography{tacl2021}
\bibliographystyle{acl_natbib}


\newpage
\appendix
\section{Appendix}
\label{sec:appendix}

\subsection{Question design for human evaluation}
\label{app:question design}
\textbf{Q1}: The rewrite deletes/obfuscates \_ ?
\begin{itemize}
    \item[(a)] None of the key information in the personal information and the original utterance does contain personal information.
    \item[(b)] None of the key information in the personal information, because the original utterance does not contain personal information.
    \item[(c)] At least one key information in the personal information (if the rewrite uses both correct and incorrect strategies, only evaluate the part that uses the correct strategy).
    \item[(d)] All key information in the personal information (using the correct strategy only). 
    \item[(e)] At least one or all key information in the personal information (using the incorrect strategies only).
\end{itemize}

\textbf{Q2}: The rewrite \_.
\begin{itemize}
    \item[(a)] Accurately preserves the meaning of the original sentence. 
    \item[(b)] Basically the same meaning but does not cover some minor content. 
    \item[(c)] Has a minor resemblance to the meaning of the original sentence, however, it is also misleading. 
    \item[(d)] Empty sentence or does not reflect the meaning of the original sentence at all.
\end{itemize}

\textbf{Q3}: The rewrite is able to retain \_ in the original utterance that is not covered in the personal information.
\begin{itemize}
         \item[(a)] has no grammatical mistakes and the sentence is coherent.
        \item[(b)] has some grammatical mistakes and the sentence is less coherent
        \item[(c)] is full of grammatical mistakes</b> and the sentence is not coherent
\end{itemize}

\subsection{\personaChat}
\label{app:personachat}
The \personaChat dataset \cite{zhang2018personaChat} is a crowd-sourced corpus designed to facilitate research in personalized open-domain dialogue systems. Each conversation in the dataset involves two speakers, each assigned a distinct persona comprising 4–5 profile sentences. These personas guide the dialogue, encouraging participants to engage in conversations that reflect their assigned characteristics.

The dataset encompasses:
\begin{itemize}
  \item \textbf{1,155} unique personas, each with at least 5 profile sentences.
  \item \textbf{10,907} dialogues totaling over \textbf{162,000} utterances.
  \item A division into training, validation, and test sets, with 100 personas reserved for validation and another 100 for testing.
\end{itemize}

This structure promotes the development of dialogue agents capable of maintaining consistent and engaging personalities throughout interactions.

\subsection{Prompt template for synethetic Data}
\label{prompt_temp}
The prompt template used across the paper is shown as \ref{prompt_template}. We use three nearest examples drawn from the training set as prompting example. Each example contains two cases if the raw persona information is provided. And objective for the prompt is to rewrite given sentence with specified strategy.
\begin{figure}[t]
\centering
\begin{tcolorbox}[
    width=\linewidth,
    colback=gray!5,
    colframe=gray!20,
    rounded corners,
    fontupper=\small
]
\textcolor{blue}{Example 1:}\\
\textcolor{orange}{If I ask you to rewrite} [\texttt{example \#1}]\\
\textcolor{green!50!black}{containing personal information} [\texttt{persona \#1}]\\
\textcolor{red}{by <deleting/obscuring> private information, you should return} [\texttt{target \#1}]\\
\textcolor{orange}{If I ask you to rewrite} [\texttt{example \#1}]\\
\textcolor{green!50!black}{containing personal information} [\texttt{empty}]\\
\textcolor{red}{by <deleting/obscuring> private information, you should return} [\texttt{example \#1}]\\[0.5em]

\textcolor{blue}{Example 2:}\\
\textcolor{orange}{If I ask you to rewrite} [\texttt{example \#2}]\\
\textcolor{green!50!black}{containing personal information} [\texttt{persona \#2}]\\
\textcolor{red}{by <deleting/obscuring> private information, you should return} [\texttt{target \#2}]\\
\textcolor{orange}{If I ask you to rewrite} [\texttt{example \#2}]\\
\textcolor{green!50!black}{containing personal information} [\texttt{empty}]\\
\textcolor{red}{by <deleting/obscuring> private information, you should return} [\texttt{example \#2}]\\[0.5em]

\textcolor{blue}{Example 3:}\\
\textcolor{orange}{If I ask you to rewrite} [\texttt{example \#3}]\\
\textcolor{green!50!black}{containing personal information} [\texttt{persona \#3}]\\
\textcolor{red}{by <deleting/obscuring> private information, you should return} [\texttt{target \#3}]\\
\textcolor{orange}{If I ask you to rewrite} [\texttt{example \#3}]\\
\textcolor{green!50!black}{containing personal information} [\texttt{empty}]\\
\textcolor{red}{by <deleting/obscuring> private information, you should return} [\texttt{example \#3}]\\[1em]

\textbf{Rewrite this sentence, deleting any private information.}\\
\textbf{Only return the rewritten sentence, nothing else.}\\
\textcolor{green!50!black}{Private information present is:} [\texttt{input persona}]\\
\textcolor{red}{Sentence to rewrite:} [\texttt{input utterance}]
\end{tcolorbox}
\caption{Prompt template for \tfive-\ourBenchmark}
\label{prompt_template}
\end{figure}

\subsection{Experiment Details}
\label{app:experiment}

\subsubsection{Evaluation metrics}

Details of the evaluation metrics for semantic relevance are provided below.
\begin{itemize}
    \item \rouge ~\cite{lin2004rouge}: It is a widely used evaluation metric measuring the overlap of unigrams between a generated text and a set of references. 
    \item \rougeLsum: It is a variant of ROUGE-L, tailored to evaluate longer texts by summarizing the longest common sub-sequences between an output text and a set of references. 
    \end{itemize}
\subsubsection{Baseline Methods}
\textbf{\dpnr.} It stands for Differentially Private Neural Representation, which applies Laplace noise to distributed representations of words in order to randomly drop sensitive words or replace sensitive words with non-sensitive ones. We compare the cosine similarity between each word in an input utterance with those in the corresponding persona, and pick the top-$k$ most similar ones. 

\textbf{\dpforward.} This method perturbs embedding matrices and multi-head attention layers during each forward pass of a language models by achieving a sentence level LDP. When adapting this approach to \tfivesmall for inference, we mainly perturb embedding matrices, because the DP mechanism for attention layers is mostly useful for protecting privacy at the training time.

\textbf{\llamaparaph.} \citet{mattern2022limits} points out the limitations of word-level LDP and propose to paraphrase input texts with lower temperature to achieve a sentence-level LDP. We implement this approach by using \llama.

\textbf{\dpprompt.}\citet{utpala2023locally} utilizes zero-shot prompting and large language model to generate document paraphrasing to prevent author de-anonymization attack which comprise the privacy of text owner with predefined utility constrain.

\textbf{\dpbart.} The method is a privatized text rewriting system incorporates LDP. The system leverages the LPD paradigm to perform model rewriting using BART model to protect input data which tackles same challenge like us.

\textbf{\flair.} we also adapt the scrubbing method used in ~\cite{lukas2023analyzing} as our baseline. We set \flair as baseline to test if the automatic method can effectively remove private information from sentences. 

\textbf{Zero-Shot LLMs.} To compare with the LLMs fine-tuned on our corpus, we apply the same prompts to the same pre-trained LLMs without any training. Specifically, we consider \tfivesmall, \llama, \chatGPT and \gptf and apply the prompt template introduced in Sec. \ref{sec:synthetic_data}. To distinguish from the fine-tuned models, the \tfivesmall and \llama in the zero-shot setting is referred to as \tfiveZero and \llamaZero, respectively.


\textbf{\tfive-\ourBenchmark.} By using the same prompts as the zero-shot version, we fine tune \tfivesmall on the training set of the manually curated corpus, with or without augmenting them with synthetic data. The prompts are similar to those used by zero-shot models detailed in \ref{prompt_temp}.

\textbf{\tfive-\ourBenchmark-DP.} To simulate the use cases that the training data of the rewriting models contains sensitive information, we apply DP-SGD~\cite{abadi2016DPSGD} when fine-tuning the \tfivesmall model in order to understand to what degree the DP mechanism impacts the inference quality of the rewriting models and shed light on future research directions.

\subsection{Implementation Details}
\label{sec:implementation}
In our experiment, we consider \tfivesmall as our targeted rewrite model, we set optimal hyperparameters for model fine tuning with learning rate of $5e^{-4}$ and beam search as decoding method with generative temperature of $0.2$. In the model finetuning, we set noise multiplier of DP-SGD~\cite{abadi2016deep} to 0.001 to gain minimal influence for model result.
In baseline experiments, for two DP methods applied to echo language model, we consider the empirically optimal noise multipliers 0.01 and epsilon to 3 with one word masked for DPNR. As for \dpforwardU, we set the key noise hyperparameters delta to $1e^{-5}$ and epsilon at $7$ to obtain the impact with small noise gap, while for \dpforwardP, we set the hyperparameters to $2e^{-5}$ and $8$ for delta and epsilon respectively.  The remaining hyperparemeters are the same as with the ones reported in the corresponding papers.

\subsection{Impact of DP-SGD.}

Table \ref{tab:RQ4_result} shows results of models trained with and without DP-SGD. The purpose is to understand to what degree the widely used DP method can influence rewriting quality if the training data is sensitive. Comparing these two settings with human rewrites, there is a slight performance drop of around 3\% with DP-SGD. However, DP-SGD provides a privacy guarantee during training which is useful when the training data is sensitive. When comparing with automatic metrics, as shown in Table \ref{synreal}, there is only a 1\% performance drop in terms of privacy leakage if DP-SGD is applied. For preservation of semantic contents, \mauve scores show little differences between using and not using DP-SGD, meaning our proposed rewriting approaches are compatible with the DP based training algorithms for more sensitive scenarios.

\begin{table}[htb]
\resizebox{0.95\columnwidth}{!}{%
\begin{tabular}{c|c|c|c}
\hline
\multicolumn{1}{l|}{} & \pStrict & \rStrict & \nStrict  \\ \hline \hline
Human\_deleting          & \textbf{82.00\%}        & 76.00\% & 95.00\% \\ \hline
DP\_deleting            & 59.00\%        & 88.00\% & 99.00\% \\ \hline
non-DP\_deleting         & 63.00\%        & 82.00\% & 96.00\% \\ \hline \hline
Human\_obscuring         & \textbf{81.00\%}       & 97.00\% & 98.00\% \\ \hline
DP\_obscuring            & 29.00\%        & 90.00\% & 98.00\% \\ \hline
non-DP\_obscuring        & 32.00\%        & 88.00\% & 93.00\% \\ \hline
\end{tabular}%
}
\caption{Human evaluation results with and without DP-SGD.}
\label{tab:RQ4_result}
\end{table}

\begin{table*}[ht]
\centering
\begin{tabular}{lrrrrrr}
\toprule
DP    & Real & Synth & LLM   & \privacyLeakage & \rouge & \rougeLsum \\
\midrule\midrule
False & 1300 & 0     & -     & 0.9190 $\pm$ 0.1077 & 0.6946 & 0.6924 \\
False & 1300 & 3900  & GPT-3 & 0.9174 $\pm$ 0.0903 & 0.7143 & 0.7122 \\
False & 1300 & 3900  & GPT-4 & \textbf{0.9381} $\pm$ \textbf{0.0870} & 0.7301 & 0.7278 \\
True  & 1300 & 0     & -     & 0.9398 $\pm$ 0.0759 & 0.7338 & 0.7316 \\
True  & 1300 & 3900  & GPT-3 & 0.9243 $\pm$ 0.0908 & 0.7368 & 0.7351 \\
True  & 1300 & 3900  & GPT-4 & 0.9297 $\pm$ 0.1135 & 0.7446 & 0.7428 \\
\bottomrule
\end{tabular}
\caption{Evaluation for DP and combination of synthetic data and human rewrites}
\label{synreal}
\end{table*}

\subsection{Private Information Alignment}\label{sparse_align}

Three alignment techniques were evaluated to determine their effectiveness in identifying utterance-persona associations: the RoBERTa MNLI entailment model, Sparse-MAX, and Sharp-MAX. The latter two algorithms, originally proposed for token-level alignment, compute an alignment matrix where each entry represents the probability that a token in an utterance leaks information about a token in a persona. Specifically, for a token $i$ in the utterance and a token $j$ in the persona, the alignment score in row $i$, column $j$ denotes the leakage likelihood.

Since our task requires sentence-level alignment rather than token-wise alignment. We modified algorithms to compute sentence-level alignment probabilities over a sample of 200 utterance-persona pairs. The goal was to correctly identify the persona associated with each utterance based on alignment scores, such that the correct utterance-persona pair receives a high score while unrelated pairs do not.

To make binary alignment decisions, a fixed threshold was applied to the alignment scores. Probabilities below the threshold were interpreted as indicating no alignment, and those above it as indicating alignment. An ideal model would achieve perfect alignment performance, with both precision and recall equal to 1.
\begin{table*}[t]
\centering
\caption{Private information alignment results.}
\label{tab:alignment_results}
\resizebox{\textwidth}{!}{%
\begin{tabular}{llrrrrrrr}
\toprule
\textbf{Family} & \textbf{Threshold} & \textbf{Recall} & \textbf{Precision} & \textbf{Min} & \textbf{Max} & \textbf{Mean} & \textbf{Frobenius Norm} & \textbf{1-Norm} \\
\midrule
\multirow{6}{*}{\textbf{RoBERTa Entailment}} 
& 0.20 & 0.71 & 0.25 & 0.00 & 0.99 & 0.04 & 12.81 & 8.08 \\
& 0.25 & 0.69 & 0.27 & 0.00 & 0.99 & 0.04 & 12.81 & 8.08 \\
& \textbf{0.30} & \textbf{0.69} & \textbf{0.28} & \textbf{0.00} & \textbf{0.99} & \textbf{0.04} & \textbf{12.81} & \textbf{8.08} \\
& 0.35 & 0.68 & 0.29 & 0.00 & 0.99 & 0.04 & 12.81 & 8.08 \\
& 0.40 & 0.68 & 0.29 & 0.00 & 0.99 & 0.04 & 12.81 & 8.08 \\
& 0.80 & 0.57 & 0.35 & 0.00 & 0.99 & 0.04 & 12.81 & 8.08 \\
\midrule
\multirow{6}{*}{\textbf{Sparse-MAX}} 
& 0.20 & 1.00 & 0.01 & 1.00 & 1.00 & 1.00 & 99.50 & 99.00 \\
& 0.25 & 1.00 & 0.01 & 1.00 & 1.00 & 1.00 & 99.50 & 99.00 \\
& 0.30 & 1.00 & 0.01 & 1.00 & 1.00 & 1.00 & 99.50 & 99.00 \\
& 0.35 & 1.00 & 0.01 & 1.00 & 1.00 & 1.00 & 99.50 & 99.00 \\
& 0.40 & 1.00 & 0.01 & 1.00 & 1.00 & 1.00 & 99.50 & 99.00 \\
& 0.80 & 1.00 & 0.01 & 1.00 & 1.00 & 1.00 & 99.50 & 99.00 \\
\midrule
\multirow{6}{*}{\textbf{Sharp-MAX}} 
& 0.20 & 1.00 & 0.01 & 0.30 & 0.43 & 0.35 & 35.77 & 38.60 \\
& 0.25 & 1.00 & 0.01 & 0.30 & 0.43 & 0.35 & 35.77 & 38.60 \\
& 0.30 & 1.00 & 0.01 & 0.30 & 0.43 & 0.35 & 35.77 & 38.60 \\
& 0.35 & 0.64 & 0.01 & 0.30 & 0.43 & 0.35 & 35.77 & 38.60 \\
& 0.40 & 0.02 & 0.03 & 0.30 & 0.43 & 0.35 & 35.77 & 38.60 \\
& 0.80 & 0.00 & NaN  & 0.30 & 0.43 & 0.35 & 35.77 & 38.60 \\
\bottomrule
\end{tabular}
}
\end{table*}
Our empirical analysis revealed that Sparse-MAX and Sharp-MAX did not generalize well to the sentence-level alignment scenario, as shown in Table.~\ref{tab:alignment_results}. This result is unsurprising given that these algorithms were originally designed for fine-grained, token-level applications. Furthermore, their performance may have been affected by the lack of hyperparameter optimization specific to the sentence-level setting. Due to time constraints, a comprehensive exploration of these configurations was not feasible. Nevertheless, future work may revisit these methods as viable options, contingent on further tuning.

In contrast, the RoBERTa-based MNLI entailment model demonstrated strong alignment performance and required minimal adaptation. Following empirical threshold analysis, a decision boundary of 0.3 was selected for determining alignment. This threshold yielded favorable results and served as the primary alignment mechanism in subsequent experiments.
\subsection{Limitation of LDP at Inference Time} 
Typical scenarios for privacy protection at inference time include i) dataset release; ii) sending queries involving sensitive queries to LLMs hosted on untrusted servers. 
Local DP can be one possible solution to adding noise locally for individual data releasing and it have more relaxed definition for user input.
LDP is designed to make local data pairs indistinguishable and work generally on a sample of instances. The mainstream LDP methods add random noise to local examples to balance privacy and utilities. The collection of modified instances are aggregated to obtain certain statistics for target tasks. The aggregation step is important to mitigate the negative effects of noise for information utility. However, privacy protection at inference time does not allow any aggregation operation among a set of instances and requires finding a tradeoff between utility and privacy for individuals. Thus LDP based text rewriting methods either add too much noise to destroy the utility of information or retain original content involving sensitive information. Our experiments demonstrate the SOTA methods based on LDP empirically and show the promising research direction using our dataset.
\subsection{Naturalness Judgment Template}
\label{append:natural_template}
We reuse the questionnaire question of naturalness to form the naturalness template for GPT-4o. We rescale the naturalness from 1-5 where 1 means very unnatural and 5 means perfectly natural. We prompt model to generate a JSON like result to score the rewritten sentence and provide the explanation on it. The detailed template is shown Figure.~\ref{fig:llm-naturalness-prompt}

\textbf{Utility test for downstream task}
To further evaluate the utility of rewrites for LLMs, we conducted additional experiments to compare the LLM's responses generated by original inputs and their rewrites as the downstream task. The original \personaChat is designed in a multiple round and chit-chat manner. We locate the input utterance in our datasets in the position of dialogue and compare the generated response with original response which can be formed as the response generation task.
Specifically, we feed original texts and their rewrites respectively to the llama3.2-3B-Instruct and compare their responses with candidate responses collected from PersonaChat dataset, from which we sampled our dataset. The candidate set contains both ground-truth and implausible responses written by human.

Generated responses were ranked by calculating the cosine similarity of their embeddings to those of ground-truth responses, using a SentenceBERT\cite{reimers2019sentence}. As shown in Table. ~\ref{tab:response_similarity}, the response similarity between the original inputs and their rewrites is 35.99\%. Both the responses generated from the original inputs and those from the rewrites achieve comparable similarity scores to the ground truth. It is also worth noting that the responses generated from the rewrites are closer to the ground truth than the implausible ones. Therefore, our rewrites achieve similar utility as original inputs.

\textbf{Generalization of Datasets}
\label{app:generalization}
To further evaluate the generalization ability of our dataset, we consider the rewriting model \bart as our backbone and fine-tune the model on our datasets. \bart is an efficient model compared to state-of-the-art language models, with only 175 million parameters. As shown in Table \ref{tab::bart}, the results demonstrate that a generative model with a small parameter size can still effectively adopt human rewrites from our dataset. The \privacyLeakage score is 85.74\%, which is competitive with GPT-3.5, and the \rougeLsum score is 57.32\%, indicating higher consistency in language compared to \llamaparaph. Considering the gap in parameter size, the results show that our dataset can enable models like \bart and \tfive to achieve competitive performance in human-like rewriting tasks.

\begin{table}
\resizebox{1\columnwidth}{!}{
\centering
\begin{tabular}{c|c|c|c}
\hline
\multicolumn{1}{l|}{} & \privacyLeakage & \rouge   & \rougeLsum \\ \hline \hline
GPT-3.5-\deleting        & 74.29\%   & 69.13\% & 68.48\%   \\ \hline

\llamaparaph-\deleting    & 76.42\%   & 56.29\% & 54.91\%   \\ \hline
\bart-\ourBenchmark   & 85.74\%   & 59.435  & 57.32\%   \\ \hline
\tfive-\ourBenchmark-\ourBenchmark & 93.81\%   & 73.01\% & 72.78\%  \\ \hline
\end{tabular}
}
\caption{\ourBenchmark finetuned \bart with baseline method} \label{tab::bart}
\end{table}


\begin{table}
\resizebox{1\columnwidth}{!}{%
\begin{tabular}{c|c|c|c}
\hline
\multicolumn{1}{l|}{} & Original&	Responses with Rewrites	&Implausible Responses \\ \hline \hline
Ground Truth Responses	  & 82.00\%        & 76.00\% & 95.00\% \\ \hline
Rewrite   & 34.00\%        & 94.00\% & 72.00\% \\ \hline

\end{tabular}%
}
\caption{Response similarity Comparison between varying types of inputs	}
\label{tab:response_similarity}
\end{table}

\begin{figure*}[t]
\centering
\begin{tcolorbox}[
    colframe=blue!50!white,
    coltitle=white,
    title=\textbf{LLM Naturalness Judgment Prompt},
    fonttitle=\bfseries,
    width=0.95\textwidth,
    boxrule=1pt,
    arc=3pt
]
You are an expert linguist. Your task is to assess the naturalness of a given sentence — how fluent, human-like, and typical it sounds in everyday language use.

\vspace{0.5em}
\textbf{Rate the sentence on a scale from 1 to 5:}
\begin{itemize}
    \item \textbf{1} = Very unnatural (full of grammatical mistakes, incoherent)
    \item \textbf{2} = Mostly unnatural
    \item \textbf{3} = Somewhat natural (acceptable, but some issues)
    \item \textbf{4} = Mostly natural (minor issues)
    \item \textbf{5} = Very natural (fluent, coherent, no errors)
\end{itemize}

\vspace{0.5em}
\textbf{Sentence:} \verb|"Sentence to Assess"|

\vspace{0.5em}
Only provide the score and a brief explanation in the following JSON format:

\begin{verbatim}
{"score": X, "explanation": "..."}
\end{verbatim}
\end{tcolorbox}
\caption{Prompt used for LLM-based naturalness judgment.}
\label{fig:llm-naturalness-prompt}
\end{figure*}

\end{document}